\documentclass[10pt,twocolumn,letterpaper]{article}

\usepackage[pagenumbers]{cvpr} % To force page numbers, e.g. for an arXiv version

\usepackage{graphicx}
\usepackage{amsmath}
\usepackage{amssymb}
\usepackage{booktabs}
\usepackage{color}
\usepackage{cases} 
\usepackage{colortbl}
\usepackage[accsupp]{axessibility}  % Improves PDF readability for those with disabilities.

\usepackage[pagebackref,breaklinks,colorlinks]{hyperref}
\usepackage{cvpr}      % To produce the REVIEW version
\usepackage{graphicx}
\usepackage{amsmath}
\usepackage{amssymb}
\usepackage{booktabs}
\usepackage{color}
\usepackage{multicol}
\usepackage{multirow}
\usepackage{array}

\usepackage[capitalize]{cleveref}
\crefname{section}{Sec.}{Secs.}
\Crefname{section}{Section}{Sections}
\Crefname{table}{Table}{Tables}
\crefname{table}{Tab.}{Tabs.}

\def\cvprPaperID{1587} % *** Enter the CVPR Paper ID here
\def\confName{CVPR}
\def\confYear{2023}
\definecolor{color3}{rgb}{0.95,0.95,0.95}

\begin{document}
\title{Optimization-Inspired Cross-Attention Transformer for Compressive Sensing}

% \and
\author{
Jiechong Song$^{1,4}$, Chong Mou$^{1}$,  Shiqi Wang$^{2}$, Siwei Ma$^{3,4}$, Jian Zhang$^{1,4 *}$\\
$^1$Peking University Shenzhen Graduate School, Shenzhen, China\\
$^2$Department of Computer Science, City University of Hong Kong, China\\
$^3$School of Computer Science, Peking University, Beijing, China\\
$^4$Peng Cheng Laboratory, Shenzhen, China\\
% {\tt\small eechongm@gmail.com; lucywq1028@gmail.com; zhangjian.sz@pku.edu.cn}
{\tt\small \{songjiechong, swma, zhangjian.sz\} @pku.edu.cn eechongm@stu.pku.edu.cn shiqwang@cityu.edu.hk}
}   
\maketitle
\let\thefootnote\relax\footnotetext{$^*$\textit{Corresponding author}. This work was supported in part by Shenzhen Research Project under Grant JCYJ20220531093215035 and Grant JSGGZD20220822095800001.}

%%%%%%%%% ABSTRACT
\begin{abstract}

By integrating certain optimization solvers with deep neural networks, deep unfolding network (DUN) with good interpretability and high performance has attracted growing attention in compressive sensing (CS). However, existing DUNs often improve the visual quality at the price of a large number of parameters and have the problem of feature information loss during iteration. In this paper, we propose an Optimization-inspired Cross-attention Transformer (OCT) module as an iterative process, leading to a lightweight \textbf{OCT}-based \textbf{U}nfolding \textbf{F}ramework (\textbf{OCTUF}) for image CS. Specifically, we design a novel Dual Cross Attention (Dual-CA) sub-module, which consists of an Inertia-Supplied Cross Attention (ISCA) block and a Projection-Guided Cross Attention (PGCA) block. ISCA block introduces multi-channel inertia forces and increases the memory effect by a cross attention mechanism between adjacent iterations. And, PGCA block achieves an enhanced information interaction, which introduces the inertia force into the gradient descent step through a cross attention block. Extensive CS experiments manifest that our OCTUF achieves superior performance compared to state-of-the-art methods while training lower complexity. Codes are available at \url{https://github.com/songjiechong/OCTUF}.

% DUNs demonstrate good interpretability and high performance but suffer from two issues. Firstly, image transmission at each iteration of traditional DUNs causes information loss. Secondly, DUNs are mainly CNN-based where the prior layers have little influence on the subsequent ones, showing limitations in capturing long-range dependencies. In this paper, we propose an Optimization-inspired Cross-attention Transformer (OCT) as each iteration of DUN for CS. 
\end{abstract}

\vspace{-10pt}

%%%%%%%%% BODY TEXT
\section{Introduction}
\label{sec:intro}

%%%%%%%%%%%%%%%%%%%%%%%%%%%%%%%%%%%%%%%%
\begin{figure}[t]
\centering
% \setlength{\abovecaptionskip}{5pt}
% \setlength{\belowcaptionskip}{-0.cm}
% \vspace{-0.3cm}
\includegraphics[width=0.95\linewidth]{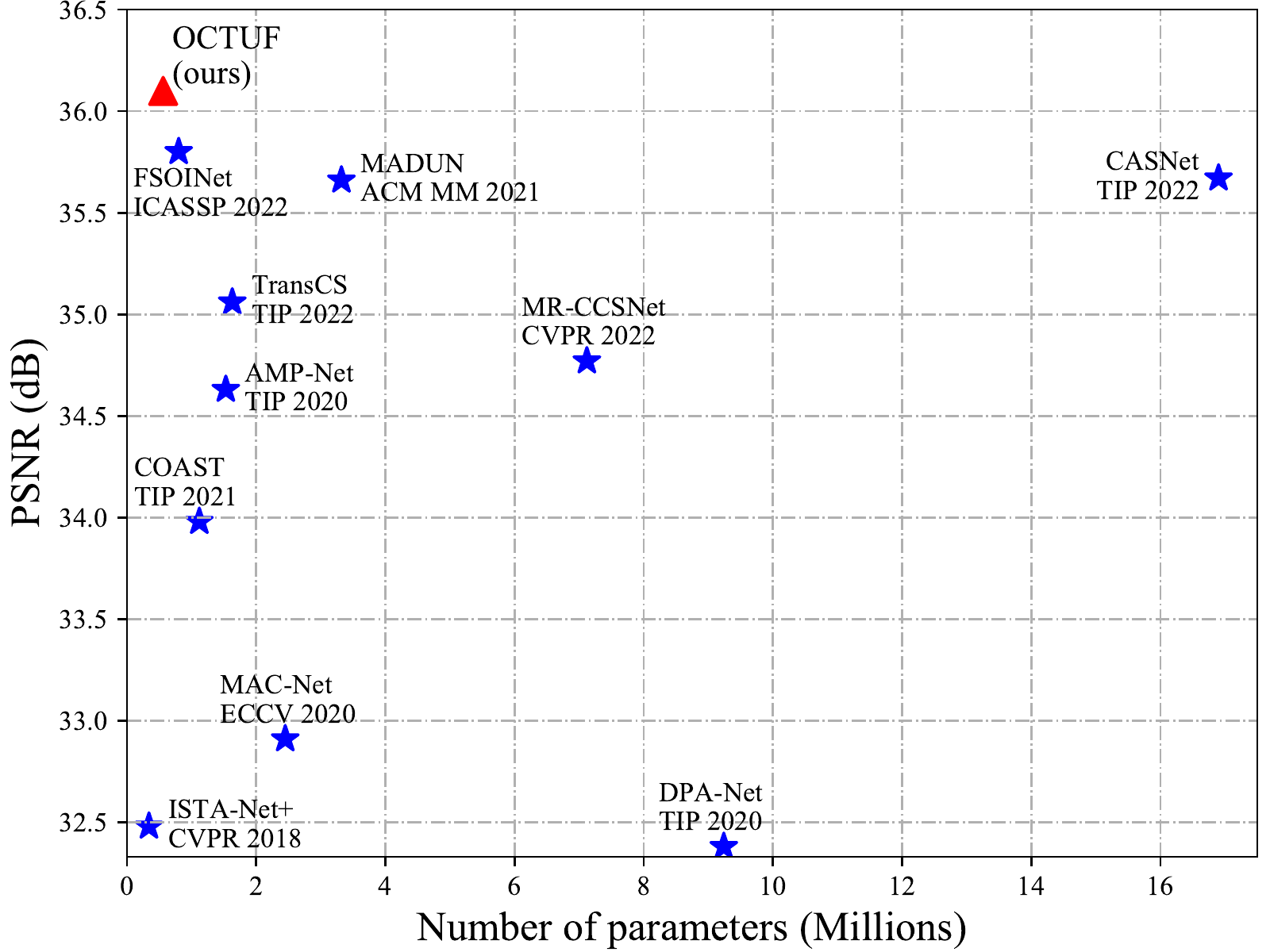} 
\vspace{-4pt}
\caption{The PSNR (dB) performance (y-axis) of our OCTUF and some recent methods (ISTA-Net \cite{zhang2018ista}, DPA-Net \cite{sun2020dual}, AMP-Net \cite{zhang2020amp}, MAC-Net \cite{josselyn2020memory}, COAST \cite{you2021coast}, MADUN \cite{song2021memory}, CASNet \cite{chen2022content}, TransCS \cite{shen2022transcs}, FSOINet \cite{chen2022fsoinet}, MR-CCSNet \cite{fan2022global}) under different parameter capacities (x-axis) on Set11 \cite{Kulkarni2016ReconNetNR} dataset in the case of CS ratio $=25\%$. Our proposed method outperforms previous methods while requiring significantly cheaper parameters.}
\label{fig:paras_psnr}
\vspace{-16pt}
\end{figure}
%%%%%%%%%%%%%%%%%%%%%%%%%%%%%%%%%%%%%%%%

Compressive sensing (CS) is a considerable research interest from signal/image processing communities as a joint acquisition and reconstruction approach \cite{candes2006robust}. The signal is first sampled and compressed simultaneously with linear random transformations. Then, the original signal can be reconstructed from far fewer measurements than that required by Nyquist sampling rate \cite{sankaranarayanan2012cs,liutkus2014imaging}. So, the two main concerns of CS are the design of the sampling matrix \cite{chen2022content,fan2022global} and recovering the original signal \cite{zhang2020amp}, and our work focuses on the latter. 
Meanwhile, the CS technology achieves great success in many image systems, including medical imaging \cite{lustig2007sparse, szczykutowicz2010dual}, 
% image compression \cite{chen2019compressive}, 
single-pixel cameras \cite{duarte2008single,rousset2016adaptive}, wireless remote monitoring \cite{zhang2012compressed}, and snapshot compressive imaging \cite{wu2021spatial,wu2021ddun,cai2022degradation}, because it can reduce the measurement and storage space while maintaining a reasonable reconstruction of the sparse or compressible signal.
% As CS can reduce the amount of information to be observed and processed while maintaining a reasonable reconstruction of the sparse or compressible signal, it has spawned many applications, including but not limited to medical imaging \cite{lustig2007sparse, szczykutowicz2010dual}, image compression \cite{chen2019compressive}, single-pixel cameras \cite{duarte2008single,rousset2016adaptive}, wireless remote monitoring \cite{zhang2012compressed}, and snapshot compressive imaging \cite{wu2021spatial,wu2021ddun,cai2022degradation}.

Mathematically, a random linear measurement $\mathbf{y}$ $\in$ $\mathbb{R}^M$ can be formulated as $\mathbf{y}$ $=$ $\mathbf{\Phi}\mathbf{x}$, where $\mathbf{x}$ $\in$ $\mathbb{R}^N$ is the original signal and $\mathbf{\Phi}$ $\in$ $\mathbb{R}^{M \times N}$ is the measurement matrix with $M$ $\ll$ $N$. $\frac{M}{N}$ is the CS ratio (or sampling rate). Obviously, CS reconstruction is an ill-posed inverse problem. To obtain a reliable reconstruction, the conventional CS methods commonly solve an energy function as:
%%%%%%%%%%%%%%%%%%%%%%%%%%%%%%%%%%%%%%%%%%
\begin{align} \label{eq: opt}
\begin{split}
\setlength{\abovedisplayskip}{5pt}
\setlength{\belowdisplayskip}{6pt}
\underset{\mathbf{x}}{\arg\min} ~~\frac{1}{2} \left \|\mathbf{\Phi}\mathbf{x} - \mathbf{y} \right \|^2_2+ \lambda \mathcal{R}(\mathbf{x}),
% \left\|\boldsymbol{\Psi\boldsymbol{\mathbf{x}}} \right\|,
\end{split}
\end{align}
%%%%%%%%%%%%%%%%%%%%%%%%%%%%%%%%%%%%%%%%%%
where $\frac{1}{2} \left \|\mathbf{\Phi}\mathbf{x} - \mathbf{y} \right \|^2_2$ denotes the data-fidelity term for modeling the likelihood of degradation and $\lambda \mathcal{R}(\mathbf{x})$ denotes the prior term with regularization parameter $\lambda$. For traditional model-based methods \cite{kim2010compressed,Li2013AnEA,zhang2014group,zhang2014image,gao2015block,Metzler2016FromDT,zhao2018cream}, the prior term can be the sparsifying operator corresponding to some pre-defined transform basis, such as discrete cosine transform (DCT) and wavelet \cite{zhao2014image,zhao2016video}. They enjoy the merits of strong convergence and theoretical analysis in most cases but are usually limited in high computational complexity and low adaptivity \cite{zhao2016nonconvex}. Recently, fueled by the powerful learning capacity of deep networks, several network-based CS algorithms have been proposed \cite{Kulkarni2016ReconNetNR,sun2020dual}.
% Apparently compared with model-based methods, 
Although network-based methods can solve CS problem adaptively
% represent image information flexibly 
with fast inferences, the architectures of most of these methods are the black box design and the advantages of traditional algorithms are not fully considered \cite{ren2021adaptive}. 

More recently, some deep unfolding networks (DUNs) with good interpretability are proposed to combine network with optimization and train a truncated unfolding inference through an end-to-end learning manner, which has become the mainstream for CS \cite{zhang2018ista,zhang2020optimization,you2021ista,you2021coast,zhang2020amp}. However, existing deep unfolding algorithms usually achieve excellent performance with a large number of iterations and a huge number of parameters \cite{song2021memory,song2023deep}, which are easily limited by storage space. Furthermore, the image-level transmission at each iteration fails to make full use of
% the correlations of 
inter-stage feature information. 
% And, existing DUNs are mainly CNN-based, therefore showing limitations in capturing non-local self-similarity and long-range dependencies, both critical for CS. 

To address the above problems, in this paper, we propose an efficient \textbf{O}ptimization-inspired \textbf{C}ross-attention \textbf{T}ransformer (\textbf{OCT}) module as the iterative process and establish a lightweight \textbf{OCT}-based \textbf{U}nfolding \textbf{F}ramework (\textbf{OCTUF}) for image CS, as shown in \cref{fig:OCT}. Our OCT module maintains maximum information flow in feature space, which consists of a Dual Cross Attention (Dual-CA) sub-module and a Feed-Forward Network (FFN) sub-module to form each iterative process. Dual-CA sub-module contains an Inertia-Supplied Cross Attention (ISCA) block and a Projection-Guided Cross Attention (PGCA) block. ISCA block calculates cross attention on adjacent iteration information and adds inertial/memory effect to the optimization algorithm. And, PGCA block uses the gradient descent step and inertial term as inputs of Cross Attention (CA) block to guide the fine fusion of channel-wise features. With the proposed techniques, OCTUF outperforms state-of-the-art CS methods with much fewer parameters, as illustrated in \cref{fig:paras_psnr}. The main contributions are summarized as follows: 
\begin{itemize}
\item
We propose a lightweight deep unfolding framework OCTUF in feature space for CS, where the optimization-inspired cross-attention Transformer (OCT) module is regarded as an iterative process.
\item
We design a compact Dual Cross Attention (Dual-CA) sub-module to guide the efficient multi-channel information interactions, which consists of a Projection-Guided Cross Attention (PGCA) block and an Inertia-Supplied Cross Attention (ISCA) block.
\item
Extensive experiments demonstrate that our proposed OCTUF outperforms existing state-of-the-art methods with cheaper computational and memory costs.
\end{itemize}

\section{Related Work}
\subsection{Deep Unfolding Network}

% Deep unfolding networks (DUNs) 
The main idea of deep unfolding networks (DUNs) is that conventional iterative optimization algorithms can be implemented equivalently by a stack of recurrent trainable blocks. Such correspondence has been proposed to solve different image inverse tasks, such as denoising \cite{chen2016trainable,lefkimmiatis2017non}, deblurring \cite{kruse2017learning,wang2020stacking}, and demosaicking \cite{kokkinos2018deep}. The solution is usually formulated as a bi-level optimization problem:
% DUN has friendly interpretability on training data pairs $\{(\mathbf{y}_j, \mathbf{x}_j)\}_{j=1}^{N_a}$, which is usually formulated on CS construction as the bi-level optimization problem:
%%%%%%%%%%%%%%%%%%%%%%%%%%%%%%%%%%%%%%%%%%
% \vspace{-3pt}
\begin{equation}
\begin{split}
&\underset{\boldsymbol{\Theta}}{\min} \sum_{j=1} {\mathcal{L}(\mathbf{\hat{x}}_j, \mathbf{x}_j)}, \\
&\mathbf{s.t.}\ \mathbf{\hat{x}}_j=\underset{\mathbf{x}}{\arg \min } \frac{1}{2} \left \| \mathbf{\Phi}\mathbf{x}-\mathbf{y}_j\right \|^2_2+\lambda\mathcal{R}(\mathbf{x}), 
\end{split}
\end{equation}
%%%%%%%%%%%%%%%%%%%%%%%%%%%%%%%%%%%%%%%%%
where $\boldsymbol{\Theta}$ denotes the trainable parameters and $\mathcal{L}(\mathbf{\hat{x}}_j, \mathbf{x}_j)$ represents the loss function of estimated clean image $\mathbf{\hat{x}}_j$ with respect to the original image $\mathbf{x}_j$. 
% Therefore, the overall network can be formulated as $\mathbf{\hat{x}} = f(\mathbf{y},\boldsymbol{\Theta})$.

% DUNs on CS
In the community of compressive sensing, DUN-based methods usually integrate some effective convolutional neural network (CNN) denoisers into some optimization methods, \textit{e.g.}, 
% including % half quadratic splitting (HQS) algorithm \cite{zhang2017learning,dong2018denoising}, 
proximal gradient descent (PGD) algorithm \cite{zhang2018ista,you2021coast,song2021memory,chenlearning,chen2022fsoinet,chen2022content,shen2022transcs}, approximate message passing (AMP) \cite{zhang2020amp}, and inertial proximal algorithm for nonconvex optimization (iPiano) \cite{su2020ipiano}. Different optimization methods lead to different optimization-inspired DUNs. In most DUNs, the input and output of each iteration are inherently images $\mathbf{x}_j$, which seriously hamper information transmission, resulting in limited representation capability \cite{zhang2020deep}. Recently, some methods \cite{chen2022fsoinet,ren2021adaptive} propose the idea of combining information flow into each iteration process in feature space to enhance information transmission. However, existing solutions usually lack flexibility in dealing with channel-wise information and are beset by high model complexity. In this paper, we present an efficient solution.
% , which extend to channel dimension directly and do not divide/integrate the differences of information in detail.

\subsection{Vision Transformer}

%%%%%%%%%%%%%%%%%%%%%%%%%%%%%%%%%%%%%%%%
\begin{figure*}[t]
\centering
\setlength{\abovecaptionskip}{0pt}
\setlength{\belowcaptionskip}{0pt}
% \vspace{-0.3cm}
\includegraphics[width=1.0\textwidth]{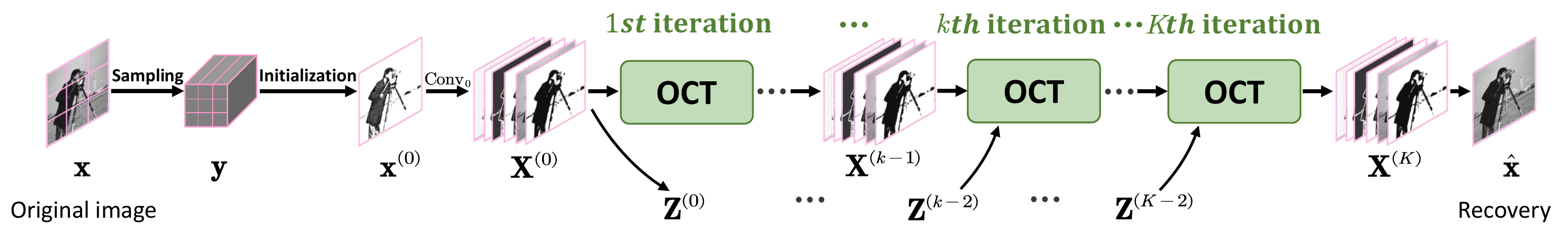} 
\vspace{-8pt}
\caption{Architecture of our OCTUF, which consists of $K$ iterations. $\mathbf{x}$ denotes the full-sampled image for training, $\mathbf{y}$ is the under-sampled data and $\mathbf{x}^{(0)}$ denotes the initialization. The feature $\mathbf{X}^{(k-1)}$ and $\mathbf{Z}^{(k-2)}$ are the inputs of our optimization-inspired cross-attention Transformer (OCT) module that is the $k$th iterative process, and $\mathbf{\hat{x}}$ is the recovered result gotten from the output $\mathbf{X}^{(K)}$ in the $K$th iteration.}
\label{fig:OCT}
\vspace{-16pt}
\end{figure*}
%%%%%%%%%%%%%%%%%%%%%%%%%%%%%%%%%%%%%%%%

Inspired by the success of Transformers  \cite{vaswani2017attention} in natural language processing, recent researchers also extend the Transformer structure for various computer vision tasks, \textit{e.g.},
% Transformer is firstly proposed by \cite{vaswani2017attention} for machine translation. Recently, Transformer has achieved great success in many vision tasks, 
% including but not limited to 
image classification \cite{dosovitskiy2020image,liu2021swin}, object detection \cite{carion2020end,zhu2020deformable}, segmentation \cite{wang2021end,petit2021u}. Transformer-based methods are also applied to image restoration tasks. PIT~\cite{pit} is the first work to introduce Transformer to image restoration and achieves promising performance in several image restoration tasks. Subsequently, several novel designs are proposed. \cite{liang2021swinir,wang2022uformer} utilize the Swin Transformer~\cite{liu2021swin} to perform image restoration.  Recently, Cai \etal \cite{cai2022degradation} propose DAUF based on DUN structure for spectral compressive imaging where self-attention is widely adopted to build the basic Transformer block and Shen \etal \cite{shen2022transcs} design an ISTA-based Transformer backbone for CS. These Transformers are just included in the prior term and have nothing to do with the data-fidelity term, so do not fully exploit the advantages of DUN. In this paper, we combine Transformer and DUN to build an efficient CS framework.
% Transformer modules are also applied for tackling the vision understanding tasks \cite{zhou2022cross, zhu2022dual}, which better learn feature embeddings for different objects.

\section{Proposed Method}
\label{sec:method}

% In this section, we will elaborate on the design of our proposed OCTUF for image compressive sensing.

\subsection{Overall Architecture}

The proximal gradient descent (PGD) algorithm is a well-suited approach for solving many large-scale linear inverse problems \cite{combettes2005signal}. Recently, some research \cite{ochs2014ipiano,boct2016inertial} finds that such an algorithm adding the inertial term always succeeds to converge to the global optimum. Ochs \etal \cite{ochs2014ipiano} propose an inertial proximal algorithm for nonconvex optimization (iPiano), which combines the gradient descent term with an inertial force. Inspired by iPiano, the whole update steps (for the $k$th iteration) can be expressed as:
% Inertial proximal algorithm for nonconvex optimization (iPiano) is an efficient but simple solver to address nonconvex issues \cite{su2020ipiano}. The CS reconstruction problem in \cref{eq: opt} can be solved by iPiano theory by iterating between the following update scheme (for the $k$th iteration):
%%%%%%%%%%%%%%%%%%%%%%%%%%%%%%%%%%%%%%%%%%
\begin{align}
\small
\label{eq:r}
&\begin{aligned}
\mathbf{s}^{(k)}=& \mathbf{x}^{(k-1)} - \rho^{(k)} \mathbf{\Phi^{\top}} (\mathbf{\Phi} \mathbf{x}^{(k-1)} - \mathbf{{y}}) \\ 
&+\alpha^{(k)} (\mathbf{x}^{(k-1)}-\mathbf{x}^{(k-2)}),
\end{aligned} \\
\label{eq:x}
&\mathbf{x}^{(k)}=\underset{\mathbf{x}}{\arg \min } \frac{1}{2}\left \|\mathbf{x}-\mathbf{s}^{(k)}\right \|^2_2+\lambda\mathcal{R}(\mathbf{x}),
\end{align}
%%%%%%%%%%%%%%%%%%%%%%%%%%%%%%%%%%%%%%%%%
where $\mathbf{x}^{(k)}$ is the output image of the $k$th iteration, $\mathbf{y}$ is the sampled image, $\rho^{(k)}, \alpha^{(k)}$ are the learnable step size parameters and $\mathbf{\Phi^{\top}}$ is the transpose of the measurement matrix $\mathbf{\Phi}$. 

\cref{eq:r} denotes the projection step, which introduces an inertial term $\alpha^{(k)} (\mathbf{x}^{(k-1)}$ $-$ $\mathbf{x}^{(k-2)} )$ to a gradient descent term $\mathcal{F}(\mathbf{x}^{(k-1)})$ $=$ $\mathbf{x}^{(k-1)}$ $-$ $\rho^{(k)} \mathbf{\Phi^{\top}} (\mathbf{\Phi} \mathbf{x}^{(k-1)}$ $-$ $\mathbf{{y}})$, relaxing the monotonically decreasing constraints and helping to achieve a better convergence result \cite{nesterov2003introductory}. As mentioned previously, such traditional implementation lacks adaptability and has information loss due to image-level inter-stage transmission. To rectify these weaknesses, we propose a \textbf{D}ual \textbf{C}ross \textbf{A}ttention (\textbf{Dual-CA}) sub-module to achieve feature-level transmission by adding a multi-channel inertial force and enhancing the information interaction in the projection step.
% increases forward information delivery. 
\cref{eq:x} is achieved by a proximal mapping step which is actually a Gaussian denoiser. Here like most DUNs, we implement it with a trainable model, \textit{i.e.}, a \textbf{F}eed \textbf{F}orward \textbf{N}etwork (\textbf{FFN}) sub-module that is detailed in \cref{fig:each_stage}(e). FFN consists of two sets of LayerNorm and Feed Forward Block (FFB) with a global skip connection, where the architecture of FFB is similar to \cite{cai2022mask}.
% , which losses the channel-wise information because the input and output are images. Motivated that multi-channel feature transmission well ensures maximum signal flow \cite{song2021memory}, we improve the whole iteration process to the feature domain, yielding the inputs of update steps are $\mathbf{X}^{(k-2)}, \mathbf{X}^{(k-1)}, \mathbf{S}^{(k)}\in\mathbb{R}^{H \times W \times C}$. Also, we expand the multi-channel inertial term for more powerful feature expression and feature correlation learning. 

Therefore, as the process in the $k$th iteration of \textbf{OCTUF}, our \textbf{O}ptimization-inspired \textbf{C}ross-attention \textbf{T}ransformer (\textbf{OCT}) module can be formulated as ($k\in\{1,2,\cdots,K\} $):
%%%%%%%%%%%%%%%%%%%%%%%%%%%%%%%%%%%%%%%%%%
\begin{align}
\small
\label{eq:dual_ca}
&\mathbf{S}^{(k)} = \mathcal{H}_{\operatorname{Dual-CA}}(\mathbf{X}^{(k-1)}, \mathbf{Z}^{(k-2)}), \\
\label{eq:pm-ffn}
&\mathbf{X}^{(k)} = \mathcal{H}_{\operatorname{FFN}}(\mathbf{S}^{(k)}),
\end{align}
%%%%%%%%%%%%%%%%%%%%%%%%%%%%%%%%%%%%%%%%%
where $\mathbf{S}^{(k)}$, $\mathbf{X}^{(k)}$ $\in$ $\mathbb{R}^{H \times W \times C}$ are the outputs in the feature domain, and $\mathbf{Z}^{(k-2)}$ $\in$ $\mathbb{R}^{H \times W \times (C-1)}$
% =\mathbf{X}^{(k-2)}[:\ ,\ :\ ,1:C]\in\mathbb{R}^{H \times W \times (C-1)}$ 
is obtained by clipping latter $C$$-1$ channels from $\mathbf{X}^{(k-2)}$. For the first iteration, the input $\mathbf{X}^{(0)}$ is generated by a $3$$\times$$3$ convolution ($\operatorname{Conv}_{0}{(\cdot)}$) on the initialization $\mathbf{x}^{(0)}$, and the inertial term is not needed \cite{ochs2014ipiano}, as shown in \cref{fig:OCT}. And the recovered result $\mathbf{\hat{x}}$ is gotten by splitting the first channel from $\mathbf{X}^{(K)}$. 

Therefore, our proposed OCTUF can skillfully integrate the inter-stage feature-level information and achieves the perfect combination with the optimization steps.
% is the part of $\mathbf{X}^{(k-2)}$ by splitting the channel dimension to reduce the computation. 
% Therefore, the iterative process in the $k$th iteration can be formulated as:
% where the forward step is regular and the inertial term can be conveniently and adaptively solved as $\mathcal{G}(\mathbf{x}^{(k-1)}, \mathbf{x}^{(k-2)}) = \alpha^{(k)} (\mathbf{x}^{(k-1)}$ $-\mathbf{x}^{(k-2)})$. 
% %%%%%%%%%%%%%%%%%%%%%%%%%%%%%%%%%%%%%%%%%%
% \begin{align}
% \small
% \label{eq:r_fd}
% &\mathbf{S}^{(k)} = \mathcal{F}(\mathbf{X}^{(k-1)}) + \mathcal{G}(\mathbf{X}^{(k-1)}, \mathbf{X}^{(k-2)}), \\
% \label{eq:x_fd}
% &\mathbf{X}^{(k)} = \mathcal{P}(\mathbf{S}^{(k)}),
% \end{align}
% %%%%%%%%%%%%%%%%%%%%%%%%%%%%%%%%%%%%%%%%%
% where $\mathbf{S}^{(k)}$, $\mathbf{X}^{(k)}$ $\in\mathbb{R}^{H \times W \times C}$ are as outputs of the $k$th iteration in feature domain and $\mathcal{F}(\cdot)$, $\mathcal{G}(\cdot)$, $\mathcal{P}(\cdot)$ represent the gradient descent term, the inertial term and the proximal model respectively. \cref{eq:x_fd} can help to better reconstruct image details by performing denoising in feature domain, however, \cref{eq:r_fd} dose not affect the performance by repeating the dimensions while increasing the computational cost. 

\subsection{Dual Cross Attention}
\label{ssec:dualca}

%%%%%%%%%%%%%%%%%%%%%%%%%%%%%%%%%%%%%%%%
\begin{figure*}[t]
\centering
\setlength{\abovecaptionskip}{0pt}
\setlength{\belowcaptionskip}{0pt}
% \vspace{-0.3cm}
\includegraphics[width=1.0\textwidth]{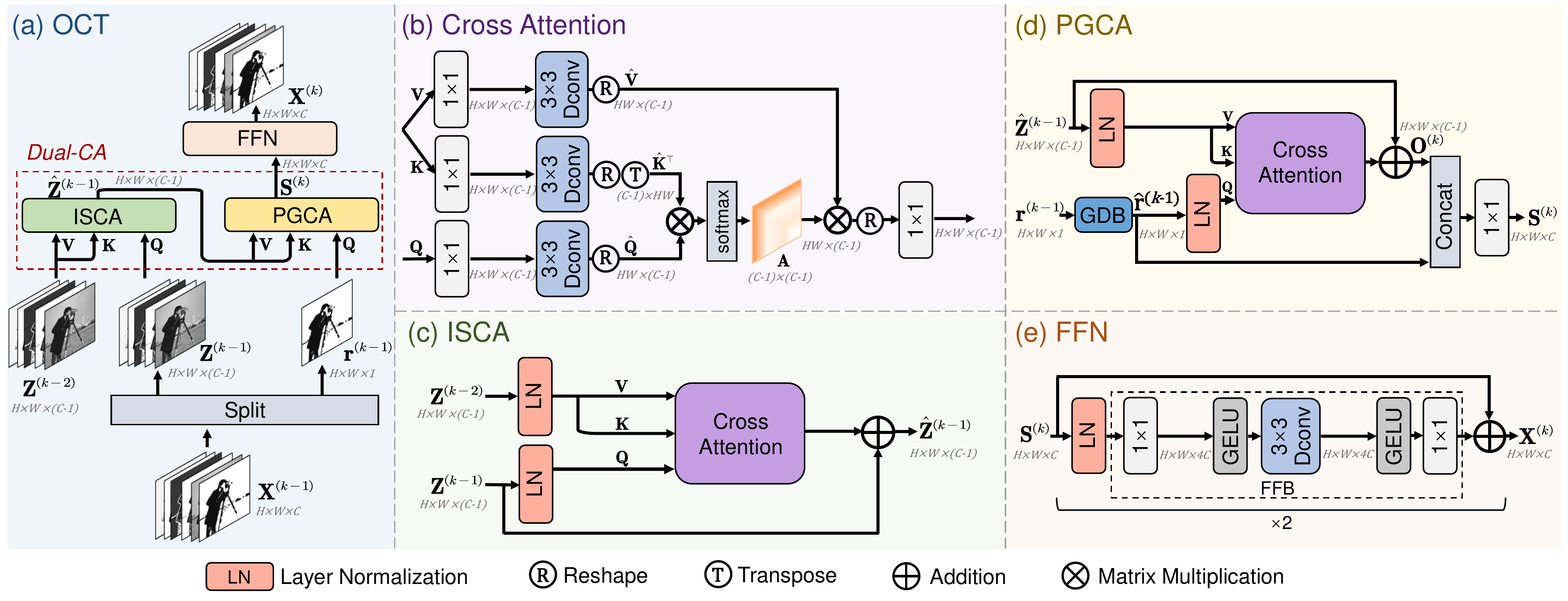} 
\vspace{-2pt}
\caption{The architecture of Optimization-inspired Cross-attention Transformer (OCT) module. (a) OCT module consists of a Dual Cross Attention (Dual-CA) sub-module which contains an Inertia-Supplied Cross Attention (ISCA) block and a Projection-Guided Cross Attention (PGCA) block, and a Feed-Forward Network (FFN) sub-module. (b) Illustration of Cross Attention (CA) block, which is the basic component of two attention blocks. (c) ISCA block is composed of Layer Normalization (LN) and CA. (d) PGCA block is composed of Gradient Descent Block (GDB), LN, and CA. (e) FFN sub-module is composed of two sets of LN and Feed-Forward Block (FFB).}
\label{fig:each_stage}
\vspace{-12pt}
\end{figure*}
%%%%%%%%%%%%%%%%%%%%%%%%%%%%%%%%%%%%%%%%

To ensure maximum information flow and powerful feature correlation, we design a Dual Cross Attention (Dual-CA) sub-module
% which consists of an Inertia-Supplied Cross Attention (ISCA) and a Projection-Guided Cross Attention (PGCA).
to efficiently fuse information in the projection step. 
As shown in \cref{fig:each_stage}(a), to make the network more compact and maintain the potential mathematical interpretation, we finely split the input $\mathbf{X}^{(k-1)}$ $\in$ $\mathbb{R}^{H \times W \times C}$ into two chunks, including $\mathbf{r}^{(k-1)}$ $\in$ $\mathbb{R}^{H \times W \times 1}$ (from the first channel) and $\mathbf{Z}^{(k-1)}\in \mathbb{R}^{H \times W \times (C-1)}$ (from the last $C$ $-$ $1$ channels).
% and then input them into
$\mathbf{r}^{(k-1)}$ and $\mathbf{Z}^{(k-1)}$ are the input of the gradient descent term and the inertial term, respectively.
% we finely split the features from the channel dimension and input them into different terms. 
% Specifically, as shown in \cref{fig:each_stage}(a), we split the input $\mathbf{X}^{(k-1)}\in\mathbb{R}^{H \times W \times C}$ into two chunks, including $\mathbf{r}^{(k-1)}\in\mathbb{R}^{H \times W \times 1}$ (from the first channel) and $\mathbf{Z}^{(k-1)}\in \mathbb{R}^{H \times W \times (C-1)}$ (from the last $C-1$ channels).
% dubbed one-channel $\mathbf{r}^{(k-1)}=\mathbf{X}^{(k-1)}[:\ ,\ :\ ,0:1]$ which is as the input of the gradient descent term, and $(C-1)$-channel $\mathbf{Z}^{(k-1)}=\mathbf{X}^{(k-1)}[:\ ,\ :\ ,1:C]$ which is as an input of the inertial term. 
So, to make full use of the information of
% to carry out a more adequate information interaction for 
the multi-channel inertial term, we design an \textbf{I}nertia-\textbf{S}upplied \textbf{C}ross \textbf{A}ttention (\textbf{ISCA}) block, yielding $\mathcal{H}_{\operatorname{ISCA}}(\mathbf{Z}^{(k-1)}, \mathbf{Z}^{(k-2)})$. And we also propose a \textbf{P}rojection-\textbf{G}uided \textbf{C}ross \textbf{A}ttention (\textbf{PGCA}) block to perform the fusion of the gradient descent term and the inertial term in a more adaptive way. 
% and break through the limitation of the addition operation. 
Therefore, our designed Dual-CA sub-module can be formulated as follows: 
% $\mathcal{H}_{\operatorname{Dual-CA}}(\mathbf{X}^{(k-1)}, \mathbf{Z}^{(k-2)})$ based on \cref{eq:dual_ca} is
%%%%%%%%%%%%%%%%%%%%%%%%%%%%%%%%%%%%%%%%%
\begin{equation}
    \mathbf{S}^{(k)} = \mathcal{H}_{\operatorname{PGCA}}(\mathcal{F}(\mathbf{r}^{(k-1)}), \mathcal{H}_{\operatorname{ISCA}}(\mathbf{Z}^{(k-1)}, \mathbf{Z}^{(k-2)})).
\end{equation}

Among ISCA and PGCA blocks, Cross Attention (CA) plays an important role as the basic block. In the following part of \cref{ssec:dualca}, we first introduce the Cross Attention block and then present ISCA and PGCA blocks, respectively.  

% \subsubsection{Cross-Attention}
\vspace{5pt}

\noindent \textbf{Cross Attention}. Motivated by modeling complex relations for generating context-aware objects in multi-modal task \cite{zhao20213dvg}, we design a Cross Attention (CA) block to aggregate the key information from the different components in the projection step of deep unfolding network, as shown in \cref{fig:each_stage}(b). The input $\mathbf{Q}$ comes from a different component than $\mathbf{V}$ and $\mathbf{K}$. They are first embedded by a $1$$\times$$1$ convolution ($\operatorname{Conv}_{\textbf{V,K,Q}}{(\cdot)}$) to obtain feature with the size being $H$$ \times$$ W$$ \times$$ (C-1)$. Then a $3$$\times$$3$ depth-wise convolution ($\operatorname{Dconv}_{\textbf{V,K,Q}}{(\cdot)}$) is used to encode channel-wise spatial context. Finally, a reshape operation ($\operatorname{R}(\cdot)$) reformulates $\mathbf{V}$, $\mathbf{K}$, and $\mathbf{Q}$ into tokens $\{ \mathbf{\hat{V}}, \mathbf{\hat{K}}, \mathbf{\hat{Q}} \}$ $\in$ $\mathbb{R}^{HW \times (C-1)}$. Therefore, this process can be defined as the following function:
%%%%%%%%%%%%%%%%%%%%%%%%%%%%%%%%%%%%%%%%%
\begin{subnumcases}{}
\mathbf{\hat{V}} = \operatorname{R}(\operatorname{Dconv}_{\mathbf{V}}(\operatorname{Conv}_{\mathbf{V}}(\mathbf{V}))), \\[4pt]
\mathbf{\hat{K}} = \operatorname{R}(\operatorname{Dconv}_{\mathbf{K}}(\operatorname{Conv}_{\mathbf{K}}(\mathbf{K}))), \\[4pt]
\label{eq:atten_q}
\mathbf{\hat{Q}} = \operatorname{R}(\operatorname{Dconv}_{\mathbf{Q}}(\operatorname{Conv}_{\mathbf{Q}}(\mathbf{Q}))).
\end{subnumcases}
%%%%%%%%%%%%%%%%%%%%%%%%%%%%%%%%%%%%%%%%%
% where $\operatorname{Conv}_{(\cdot)}$ is the $1 \times 1$ point-wise convolution, $\operatorname{Dconv}_{(\cdot)}$ is the $3 \times 3$ depth-wise convolution and $\operatorname{R}$ is the reshape operation. 
Next, a transposed attention map $\mathbf{A}$ $\in$ $\mathbb{R}^{(C-1) \times (C-1)}$ is generated by applying softmax function to re-weight the matrix multiplication $\mathbf{\hat{K}^\top}\mathbf{\hat{Q}}$, yielding
%%%%%%%%%%%%%%%%%%%%%%%%%%%%%%%%%%%%%%%%%%
\begin{equation}
\mathbf{A} = \operatorname{Softmax}(\mathbf{\hat{K}^\top}\mathbf{\hat{Q}}),
\end{equation}
%%%%%%%%%%%%%%%%%%%%%%%%%%%%%%%%%%%%%%%%%%
where $\mathbf{\hat{K}^\top}$ denotes the transposed matrix of $\mathbf{\hat{K}}$. The aggregation result is calculated as $\mathbf{\hat{V}}\mathbf{A}$, which is reshaped into the features of size $\mathbb{R}^{H \times W \times (C-1)}$. Finally, we apply a $1$$\times$$1$ convolution $\operatorname{Conv}_{\mathbf{A}}(\cdot)$ to enhance the feature extraction. Overall, the Cross Attention block is defined as:
%%%%%%%%%%%%%%%%%%%%%%%%%%%%%%%%%%%%%%%%%%
\begin{equation}
\mathcal{G}_{\operatorname{CA}}(\mathbf{V}, \mathbf{K}, \mathbf{Q}) = \operatorname{Conv}_{\mathbf{A}}(\operatorname{R}(\mathbf{\hat{V}}\mathbf{A})).
\end{equation}
%%%%%%%%%%%%%%%%%%%%%%%%%%%%%%%%%%%%%%%%%%
Cross Attention block helps to extract useful information via 
% between the different branches by the 
channel-wise similarity with low computational cost.

%%%%%%%%%%%%%%%%%%%%%%%%%%%%%%%%%%%%
\begin{table*}[t]
		\centering
% 		\setlength{\abovecaptionskip}{0pt}
        % \setlength{\belowcaptionskip}{1pt}
% 		\small
		\caption{Average PSNR(dB)/SSIM performance comparisons of recent deep network-based CS methods on Set11 dataset \cite{Kulkarni2016ReconNetNR} with different CS ratios. The best and second-best results are highlighted in \textcolor{red}{red} and \textcolor{blue}{blue} colors, respectively.}
        \vspace{-5pt}
		\label{tab:set11}
% 		\vspace{5pt}
		\resizebox{\textwidth}{!}{%
		\renewcommand\tabcolsep{5pt}
			\begin{tabular}{c|c|cccccc}
			\toprule
				% \hline
				\multicolumn{1}{c|}{\multirow{2}{*}{Dataset}}&\multicolumn{1}{c|}{\multirow{2}{*}{Methods}}&\multicolumn{6}{c}{CS Ratio}\\ 
				\cline{3-8}
				\multicolumn{1}{c|}{} &  \multicolumn{1}{c|}{}  &\multicolumn{1}{c}{10\%}&\multicolumn{1}{c}{25\%}&\multicolumn{1}{c}{30\%}&\multicolumn{1}{c}{40\%}&\multicolumn{1}{c}{50\%}&\multicolumn{1}{c}{Average}\\ 
				\toprule % \hline% \hline
				\multirow{10}{*}{Set11}
				&\multicolumn{1}{l|}{ISTA-Net$^+$ (CVPR 2018) \cite{zhang2018ista}} &26.58/0.8066&32.48/0.9242&33.81/0.9393&36.04/0.9581&38.06/0.9706&33.39/0.9197\\
				&\multicolumn{1}{l|}{DPA-Net (TIP 2020) \cite{sun2020dual}} &27.66/0.8530&32.38/0.9311&33.35/0.9425&35.21/0.9580&36.80/0.9685&33.08/0.9306\\
				&\multicolumn{1}{l|}{AMP-Net (TIP 2020) \cite{zhang2020amp}} &29.40/0.8779&34.63/0.9481&36.03/0.9586&38.28/0.9715&40.34/0.9804&35.74/0.9473\\ 
				&\multicolumn{1}{l|}{MAC-Net (ECCV 2020) \cite{chenlearning}} &27.68/0.8182&32.91/0.9244&33.96/0.9372&35.94/0.9560&37.67/0.9668&33.63/0.9205\\ 
				&\multicolumn{1}{l|}{COAST (TIP 2021) \cite{you2021coast}} &28.74/0.8619&33.98/0.9407&35.11/0.9505&37.11/0.9646&38.94/0.9744&34.78/0.9384\\ 
				&\multicolumn{1}{l|}{MADUN (ACM MM 2021) \cite{song2021memory}} &29.91/0.8986&35.66/0.9601&36.94/\textcolor{red}{0.9676}&39.15/0.9772&40.77/\textcolor{blue}{0.9832}&36.48/0.9573\\
				&\multicolumn{1}{l|}{CASNet (TIP 2022) \cite{chen2022content}} &30.36/0.9014&35.67/0.9591&36.92/0.9662&39.04/0.9760&40.93/0.9826&36.58/0.9571\\
                &\multicolumn{1}{l|}{TransCS (TIP 2022) \cite{shen2022transcs}} 
                &29.54/0.8877&35.06/0.9548&35.62/0.9588&38.46/0.9737&40.49/0.9815&35.83/0.9513\\
                &\multicolumn{1}{l|}{FSOINet (ICASSP 2022) \cite{chen2022fsoinet}} &30.46/0.9023&\textcolor{blue}{35.80}/0.9595&37.00/0.9665&39.14/0.9764&41.08/\textcolor{blue}{0.9832}&36.70/0.9576\\
				&\multicolumn{1}{l|}{MR-CCSNet (CVPR 2022) \cite{fan2022global}} &-/-&34.77/0.9546&-/-&-/-&40.73/0.9828&-/-\\
				\cline{2-8}
				&\multicolumn{1}{l|}{OCTUF (Ours)} &\textcolor{blue}{30.70}/\textcolor{blue}{0.9030}&\textcolor{red}{36.10}/\textcolor{blue}{0.9604}&\textcolor{blue}{37.21}/\textcolor{blue}{0.9673}&\textcolor{blue}{39.41}/\textcolor{blue}{0.9773}&\textcolor{blue}{41.34}/\textcolor{red}{0.9838}&\textcolor{blue}{36.95}/\textcolor{blue}{0.9584}\\
				&\multicolumn{1}{l|}{OCTUF$^+$ (Ours)} &\textcolor{red}{30.73}/\textcolor{red}{0.9036}&\textcolor{red}{36.10}/\textcolor{red}{0.9607}&\textcolor{red}{37.32}/\textcolor{red}{0.9676}&\textcolor{red}{39.43}/\textcolor{red}{0.9774}&\textcolor{red}{41.35}/\textcolor{red}{0.9838}&\textcolor{red}{36.99}/\textcolor{red}{0.9586}\\
				\toprule
		\end{tabular}}
		\vspace{-12pt}
	\end{table*}
%%%%%%%%%%%%%%%%%%%%%%%%%%%%%%%%%%%%

% \subsubsection{Inertia-Supplied Cross-channel Attention}
\vspace{5pt}
\noindent \textbf{Inertia-Supplied Cross Attention}. As shown in \cref{eq:r}, the general inertial term usually adopts the simple operation by directly subtracting the adjacent iteration output, which is proved to be ineffective in the ablation of \cref{tab:attention}. To enrich the information interaction of the inertial term, we introduce a multi-channel inertial term and propose an Inertia-Supplied Cross Attention (ISCA) block. Our ISCA block consists of LayerNorm (LN) function and CA block as shown in \cref{fig:each_stage}(c). Specifically, we set the $(k$$-$$2)$th iteration output $\mathbf{Z}^{(k-2)}$ as \textit{value} ($\mathbf{V}^{(k)}_{\operatorname{ISCA}}$) and \textit{key} ($\mathbf{K}^{(k)}_{\operatorname{ISCA}}$), and we set the $(k$$-$$1)$th iteration output $\mathbf{Z}^{(k-1)}$ as \textit{query} ($\mathbf{Q}^{(k)}_{\operatorname{ISCA}}$), pass through CA block after normalization by LN function, so 
% ISCA block can calculate 
$\mathbf{\hat{Z}}^{(k-1)}=\mathcal{H}_{\operatorname{ISCA}}(\mathbf{Z}^{(k-1)}, \mathbf{Z}^{(k-2)})$ as:
% be formulated $\mathbf{\hat{Z}}^{(k-1)}=\mathcal{H}_{\operatorname{ISCA}}(\mathbf{Z}^{(k-1)},\mathbf{Z}^{(k-2)})$ as:
%%%%%%%%%%%%%%%%%%%%%%%%%%%%%%%%%%%%%%%%%%
% \begin{small}
\begin{equation}
\begin{split}
    &\mathbf{V}^{(k)}_{\operatorname{ISCA}},\ \mathbf{K}^{(k)}_{\operatorname{ISCA}},\ \mathbf{Q}^{(k)}_{\operatorname{ISCA}} =\\
    &\ \ \ \ \ \ \ \ \ \ \ \operatorname{LN}(\mathbf{Z}^{(k-2)}),\ \operatorname{LN}(\mathbf{Z}^{(k-2)}),\ \operatorname{LN}(\mathbf{Z}^{(k-1)}),\\ 
    % \mathbf{K}^{(k)}_{\operatorname{ISCA}} =& \operatorname{LayerNorm}(\mathbf{Z}^{(k-2)}),\\
    % \mathbf{Q}^{(k)}_{\operatorname{ISCA}} =& \operatorname{LayerNorm}(\mathbf{Z}^{(k-1)}),\\
    &\mathbf{\hat{Z}}^{(k-1)} = \mathcal{G}_{\operatorname{CA}}(\mathbf{V}^{(k)}_{\operatorname{ISCA}}, \mathbf{K}^{(k)}_{\operatorname{ISCA}}, \mathbf{Q}^{(k)}_{\operatorname{ISCA}}) + \mathbf{Z}^{(k-1)}.
\end{split}
\end{equation}
% \end{small}
%%%%%%%%%%%%%%%%%%%%%%%%%%%%%%%%%%%%%%%%%%
ISCA block adaptively learns more useful multi-channel inertial force and enhances memory effect to our network.
\vspace{5pt}

% \subsubsection{Projection-Guided Cross-channel Attention}
\noindent \textbf{Projection-Guided Cross Attention}. In order to adaptively combine the gradient descent term and the inertial term,
% make the network more compact and reduce the computational cost of the inter-channel redundancy
we introduce a Projection-Guided Cross Attention (PGCA) block in \cref{fig:each_stage}(d). Similar to ISCA block, PGCA block captures rich feature information based on channel-wise similarity.
% which can reasonably represent different items of the projection module by channel-wise division in detail. 
Specifically, given $\mathbf{X}^{(k-1)}$, the input of gradient descent term is gotten by its first channel (\textit{i.e.}, $\mathbf{r}^{(k-1)}$). So, the calculation of the term has the following expression:
% Mathematically, the gradient descent term is   
% is   is separated from the output  of $(k-1)$th iteration as the input of the gradient descent block (GDB) to perform the forward term, namely 
%%%%%%%%%%%%%%%%%%%%%%%%%%%%%%%%%%%%%%%%%%
\begin{equation}
\mathbf{\hat{r}}^{(k-1)}= \mathbf{r}^{(k-1)} - \rho^{(k)} \mathbf{\Phi^{\top}} (\mathbf{\Phi} \mathbf{r}^{(k-1)} - \mathbf{{y}}).
\end{equation}
%%%%%%%%%%%%%%%%%%%%%%%%%%%%%%%%%%%%%%%%%
Next, $\hat{\mathbf{r}}^{(k-1)}$ and the ISCA output $\hat{\mathbf{Z}}^{(k-1)}$ pass through LayerNorm function and CA block, yielding
% Then a position embedding, which can be expressed as:
%%%%%%%%%%%%%%%%%%%%%%%%%%%%%%%%%%%%%%%%%%
% \begin{small}
\begin{equation}
\begin{split}
    &\mathbf{V}^{(k)}_{\operatorname{PGCA}},\ \mathbf{K}^{(k)}_{\operatorname{PGCA}},\ \mathbf{Q}^{(k)}_{\operatorname{PGCA}} =\\
    &\ \ \ \ \ \ \ \ \ \ \ \operatorname{LN}(\mathbf{\hat{Z}}^{(k-1)}),\operatorname{LN}(\mathbf{\hat{Z}}^{(k-1)}),\operatorname{LN}(\mathbf{\hat{r}}^{(k-1)}),\\ 
    % \mathbf{V}^{(k)}_{\operatorname{PGCA}} =& \operatorname{LayerNorm}(\mathbf{\hat{Z}}^{(k-1)}),\\ 
    % \mathbf{K}^{(k)}_{\operatorname{PGCA}} =& \operatorname{LayerNorm}(\mathbf{\hat{Z}}^{(k-1)}),\\
    % \mathbf{Q}^{(k)}_{\operatorname{PGCA}} =& \operatorname{LayerNorm}(\mathbf{\hat{r}}^{(k-1)}),\\
    &\mathbf{O}^{(k)} = \mathcal{G}_{\operatorname{CA}}(\mathbf{V}^{(k)}_{\operatorname{PGCA}}, \mathbf{K}^{(k)}_{\operatorname{PGCA}}, \mathbf{Q}^{(k)}_{\operatorname{PGCA}})
    % + \operatorname{PE}(\mathbf{V}^{(k)}_{\operatorname{PGCA}}) 
    + \mathbf{\hat{Z}}^{(k-1)}.
\end{split}
\end{equation}
% \end{small}
%%%%%%%%%%%%%%%%%%%%%%%%%%%%%%%%%%%%%%%%%%
% where $\operatorname{PE}(\cdot)$ is the function to generate position embedding. It consists of two $3 \times 3$ depth-wise convolution layers and a GELU activation. 
It is worth noting that $\mathbf{\hat{r}}^{(k-1)}$ generates feature maps with $C-1$ channels by $\operatorname{Conv}_{\mathbf{Q}}(\cdot)$ of \cref{eq:atten_q} to enrich the feature expression. Finally, $\mathbf{O}^{(k)}$ and $\mathbf{\hat{r}}^{(k-1)}$ are concatenated, reshaped to match the original  channel dimensions and mixed with a $1$$\times$$1$ convolution ($\operatorname{Conv}_{\mathbf{O}}(\cdot)$):
% . Therefore, the process of PGCA can be reformulated as:
% PGCA calculates $\mathbf{S}^{(k)}$ as:
% an be formulated $\mathbf{S}^{(k)}=\mathcal{H}_{\operatorname{PGCA}}(\mathbf{\hat{Z}}^{(k-1)},\mathbf{\hat{r}}^{(k-1)})$ as
%%%%%%%%%%%%%%%%%%%%%%%%%%%%%%%%%%%%%%%%%%
\begin{equation}
\mathbf{S}^{(k)}= \operatorname{Conv}_{\mathbf{O}}(\operatorname{Concat}(\mathbf{O}^{(k)},\mathbf{\hat{r}}^{(k-1)})).
\end{equation}
%%%%%%%%%%%%%%%%%%%%%%%%%%%%%%%%%%%%%%%%%%
PGCA not only inherits the advantage of \cref{eq:r} but also tactfully achieves the multi-channel feature fusion between the gradient descent term and the inertial term.

% \subsection{Proximal-Mapping Feed-Forward Network}
% \label{ssec:pmffn}

\subsection{Loss Function}
\label{ssec:loss}

%%%%%%%%%%%%%%%%%%%%%%%%%%%%%%%%%%%%%%%%
\begin{figure*}[t]
\centering
% \setlength{\abovecaptionskip}{5pt}
% \setlength{\belowcaptionskip}{-0.cm}
% \vspace{-0.3cm}
\includegraphics[width=1.0\textwidth]{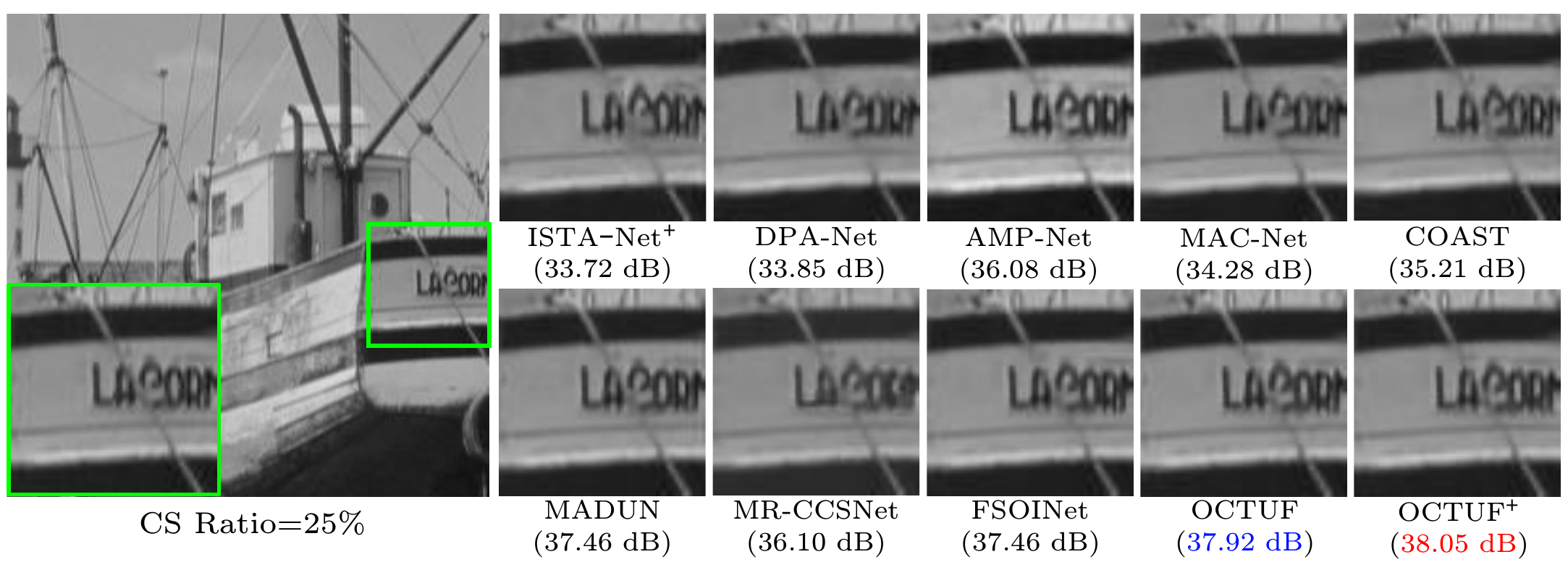}
\vspace{-20pt}
\caption{Comparisons on recovering an image from Set11 dataset \cite{Kulkarni2016ReconNetNR} in the case of CS ratio $=25 \%$. 
% The proposed OCTUFs better preserve fine structural patterns in the image.
}
\label{fig:set11}
\vspace{-12pt}
\end{figure*}
%%%%%%%%%%%%%%%%%%%%%%%%%%%%%%%%%%%%%%%%
%%%%%%%%%%%%%%%%%%%%%%%%%%%%%%%%%%%%%%%%
\begin{figure*}[t]
\centering
% \setlength{\abovecaptionskip}{5pt}
% \setlength{\belowcaptionskip}{-0.cm}
% \vspace{-0.3cm}
\includegraphics[width=1.0\textwidth]{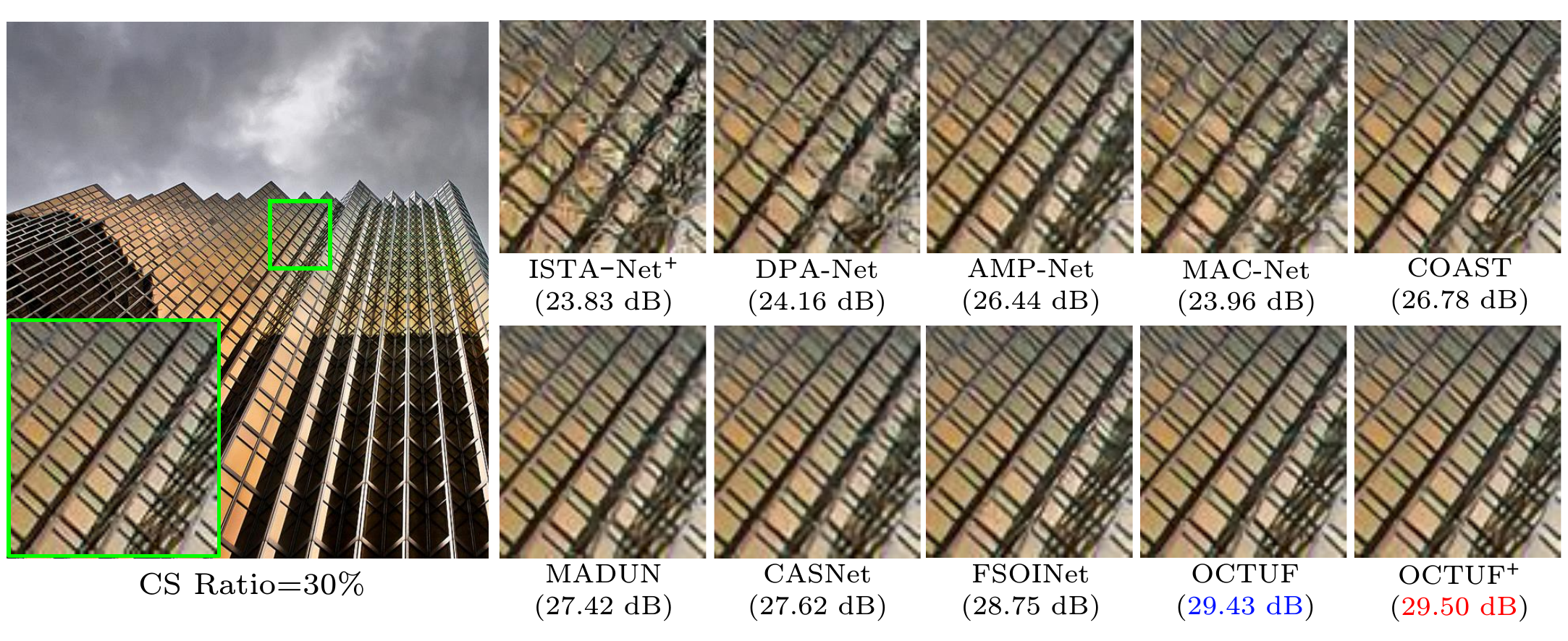} 
\vspace{-20pt}
\caption{Comparisons on recovering an image from Urban100 dataset \cite{dong2018denoising} in the case of CS ratio $=30 \%$. 
% Our OCTUFs are able to recover fine details and leave a few artifacts.
}
\label{fig:urban100}
\vspace{-12pt}
\end{figure*}
%%%%%%%%%%%%%%%%%%%%%%%%%%%%%%%%%%%%%%%%

Given a set of full-sampled images $\left\{\mathbf{x}_j\right\}_{j=1}^{N_a}$ and some sampling patterns with the specific sampling rate, the compressed measurements can be obtained by $\mathbf{y}_j=\mathbf{\Phi}\mathbf{x}_j$, producing the train data pairs $\{(\mathbf{y}_j, \mathbf{x}_j)\}_{j=1}^{N_a}$. Our model takes $\mathbf{y}_j$ as input and generates the reconstruction result $\mathbf{\hat{x}}_j$ as output. We employ the MSE loss function with respect to $\mathbf{x}_j$ and $\mathbf{\hat{x}}_j$ as following shows:
\begin{equation} 
\setlength{\abovedisplayskip}{5pt}
\setlength{\belowdisplayskip}{6pt}
\label{eq: fifteen}
\mathcal{L}(\boldsymbol{\Theta}) =\ \frac{1}{{N}{N_a}}\sum_{j=1}^{N_a} {\left\|\mathbf{x}_j-\mathbf{\hat{x}}_j\right\|}^2_2, 
\end{equation}
where $N_a$ and $N$ represent the number of the training images and the size of each image respectively. $\boldsymbol{\Theta}$ denotes the learnable parameter set of our proposed OCTUF and can be formulated as $\boldsymbol{\Theta}=\{\mathbf{\Phi}, \operatorname{Conv}_{0}(\cdot)\}\bigcup\{\mathcal{H}_{\operatorname{Dual-CA}}^{(k)}(\cdot), \mathcal{H}_{\operatorname{FFN}}^{(k)}(\cdot)\}_{k=1}^{K}$. 

\section{Experiments}

\subsection{Implementaion Details}
\label{ssec:exper_detail}

For training, we use 400 images from the training and test dataset of BSD500 dataset \cite{arbelaez2010contour}. The training images are cropped to 89600 patches of $96$$\times$$96$ pixel size with data augmentation following \cite{shi2019image}. For a given CS ratio $\frac{M}{N}$, the corresponding learnable measurement matrix $\mathbf{\Phi}$ is constructed by a convolution layer with the kernel size of $M$$ \times 1 $$\times \sqrt{N} $$\times$$\sqrt{N}$ to sample the original image $\mathbf{x}$. And then, we utilize the transpose convolution whose kernel weight is the sampling matrix to obtain initialization $\mathbf{x}^{(0)}$.

For the network parameters, the block size is 32, \textit{i.e.} $N=1,024$, the default batch size is 16, the default number of feature maps $C$ is 32 and the learnable parameter $\rho^{(k)}$ is initialized to 0.5. We use Adam \cite{Kingma2015AdamAM} optimizer to train the network with the initial learning rate, which decreased to $5$$\times 10^{-5}$ through 100 epochs using the cosine annealing strategy \cite{loshchilov2016sgdr, chen2022fsoinet} and the warm-up epochs are 3. For testing, we utilize two widely-used benchmark datasets, yielding Set11 \cite{Kulkarni2016ReconNetNR} and Urban100 \cite{dong2018denoising}. Color images are processed in the YCbCr space and evaluated on the Y channel. Two common-used image assessment criteria, Peak Signal to Noise Ratio (PSNR) and Structural Similarity (SSIM), are adopted to evaluate the reconstruction results. We also show the number of parameters and the computation cost (including the computations of the convolution, the fully-connected layer and matrix multiplication) measured in floating-point operations per second (FLOPs). 

%%%%%%%%%%%%%%%%%%%%%%%%%%%%%%%%%%%%
\begin{table*}[t]
		\centering
% 		\setlength{\abovecaptionskip}{0pt}
        % \setlength{\belowcaptionskip}{1pt}
% 		\small
		\caption{Average PSNR(dB)/SSIM performance comparisons of recent deep network-based CS methods on Urban100 dataset \cite{dong2018denoising} with different CS ratios. The best and second best results are highlighted in \textcolor{red}{red} and \textcolor{blue}{blue} colors, respectively.}
		\label{tab:urban100}
		\vspace{-3pt}
		\resizebox{\textwidth}{!}{%
		\renewcommand\tabcolsep{5pt}
			\begin{tabular}{c|c|cccccc}
			\toprule
				% \hline
				\multicolumn{1}{c|}{\multirow{2}{*}{Dataset}}&\multicolumn{1}{c|}{\multirow{2}{*}{Methods}}&\multicolumn{6}{c}{CS Ratio}\\ 
				\cline{3-8}
				\multicolumn{1}{c|}{} &  \multicolumn{1}{c|}{}  &\multicolumn{1}{c}{10\%}&\multicolumn{1}{c}{25\%}&\multicolumn{1}{c}{30\%}&\multicolumn{1}{c}{40\%}&\multicolumn{1}{c}{50\%}&\multicolumn{1}{c}{Average}\\ 
				\toprule % \hline% \hline
				\multirow{10}{*}{Urban100}
				&\multicolumn{1}{l|}{ISTA-Net$^+$ (CVPR 2018) \cite{zhang2018ista}} &23.61/0.7238&28.93/0.8840&30.21/0.9079&32.43/0.9377&34.43/0.9571&29.92/0.8821\\
				&\multicolumn{1}{l|}{DPA-Net (TIP 2020) \cite{sun2020dual}} &24.55/0.7841&28.80/0.8944&29.47/0.9034&31.09/0.9311&32.08/0.9447&29.20/0.8915\\
				&\multicolumn{1}{l|}{AMP-Net (TIP 2020) \cite{zhang2020amp}} &26.04/0.8151&30.89/0.9202&32.19/0.9365&34.37/0.9578&36.33/0.9712&31.96/0.9202\\ 
				&\multicolumn{1}{l|}{MAC-Net (ECCV 2020) \cite{chenlearning}} &24.21/0.7445&28.79/0.8798&29.99/0.9017&31.94/0.9272&34.03/0.9513&29.79/0.8809\\ 
				&\multicolumn{1}{l|}{COAST (TIP 2021) \cite{you2021coast}} &25.94/0.8035&31.10/0.9168&32.23/0.9321&34.22/0.9530&35.99/0.9665&31.90/0.9144\\ 
				&\multicolumn{1}{l|}{MADUN (ACM MM 2021) \cite{song2021memory}} &27.13/0.8393&32.54/0.9347&33.77/0.9472&35.80/0.9633&37.75/0.9746&33.40/0.9318\\
				&\multicolumn{1}{l|}{CASNet (TIP 2022) \cite{chen2022content}} &27.46/0.8616&32.20/0.9396&33.37/0.9511&35.48/0.9669&37.45/0.9777&33.19/0.9394\\
                &\multicolumn{1}{l|}{TransCS (TIP 2022) \cite{shen2022transcs}} &26.72/0.8413&31.72/0.9330&31.95/0.9483&35.22/0.9648&37.20/0.9761&32.56/0.9327\\
				&\multicolumn{1}{l|}{FSOINet (ICASSP 2022) \cite{chen2022fsoinet}} &27.53/\textcolor{blue}{0.8627}&32.62/0.9430&33.84/0.9540&35.93/\textcolor{blue}{0.9688}&37.80/0.9777&33.54/0.9412\\ \cline{2-8}
				&\multicolumn{1}{l|}{OCTUF (Ours)} &\textcolor{blue}{27.79}/0.8621&\textcolor{blue}{32.99}/\textcolor{blue}{0.9445}&\textcolor{blue}{34.21}/\textcolor{blue}{0.9555}&\textcolor{blue}{36.25}/0.9669&\textcolor{red}{38.29}/\textcolor{red}{0.9797}&\textcolor{blue}{33.91}/\textcolor{blue}{0.9423}\\
				&\multicolumn{1}{l|}{OCTUF$^+$ (Ours)} &\textcolor{red}{27.92}/\textcolor{red}{0.8652}&\textcolor{red}{33.08}/\textcolor{red}{0.9453}&\textcolor{red}{34.27}/\textcolor{red}{0.9559}&\textcolor{red}{36.31}/\textcolor{red}{0.9700}&\textcolor{blue}{38.28}/\textcolor{blue}{0.9795}&\textcolor{red}{33.97}/\textcolor{red}{0.9432}\\
				\toprule
		\end{tabular}}
\vspace{-6pt}
	\end{table*}
%%%%%%%%%%%%%%%%%%%%%%%%%%%%%%%%%%%%
%%%%%%%%%%%88%%%%%%%%%%%%%%%%%%
\begin{table*}[t]
\centering
% \small
% \normalsize
%\footnotesize 
\caption{Ablation study of our approach on Set11 dataset \cite{Kulkarni2016ReconNetNR} in the case of CS ratio $=50\%$. The best performance is labeled in bold.} 
\vspace{-3pt}
\label{tab:ablation study}
\begin{tabular}{c|c c c|c|c|c|c}
% \hline
\toprule[1pt]
\rowcolor{color3} \multicolumn{1}{c|}{\multirow{1}{*}{Cases}} & \multicolumn{1}{c}{\multirow{1}{*}{Dual-CA}} & \multicolumn{1}{c}{\multirow{1}{*}{FFN}} & \multicolumn{1}{c|}{\multirow{1}{*}{LayerNorm}} & \multicolumn{1}{c|}{\multirow{1}{*}{Learning rate}} & \multicolumn{1}{c|}{\multirow{1}{*}{PSNR(dB)}} & \multicolumn{1}{c|}{\multirow{1}{*}{SSIM}} & \multicolumn{1}{c}{\multirow{1}{*}{Parameters}}
\\ \hline  
\multicolumn{1}{c|}{(a)} & \multicolumn{1}{c}{-} & \multicolumn{1}{c}{-} & \multicolumn{1}{c|}{-} & \multicolumn{1}{c|}{5e-4(warmup)} & \multicolumn{1}{c|}{38.25} & \multicolumn{1}{c}{0.9759} & \multicolumn{1}{|c}{0.72 M}
\\  
\multicolumn{1}{c|}{(b)} & \multicolumn{1}{c}{-} & \multicolumn{1}{c}{$\surd$} & \multicolumn{1}{c|}{$\surd$} & \multicolumn{1}{c|}{5e-4(warmup)} & \multicolumn{1}{c|}{38.96} & \multicolumn{1}{c}{0.9783} & \multicolumn{1}{|c}{0.72 M} 
\\
\multicolumn{1}{c|}{(c)} & \multicolumn{1}{c}{$\surd$} & \multicolumn{1}{c}{-} & \multicolumn{1}{c|}{$\surd$} & \multicolumn{1}{c|}{5e-4(warmup)} & \multicolumn{1}{c|}{41.16} & \multicolumn{1}{c}{0.9834} & \multicolumn{1}{|c}{0.82 M} 
\\ 
\multicolumn{1}{c|}{(d)} & \multicolumn{1}{c}{$\surd$} & \multicolumn{1}{c}{$\surd$} & \multicolumn{1}{c|}{-} & \multicolumn{1}{l|}{5e-4(warmup)} & \multicolumn{1}{c|}{41.21} & \multicolumn{1}{c}{0.9834} & \multicolumn{1}{|c}{0.82 M}
\\
\multicolumn{1}{c|}{(e)} & \multicolumn{1}{c}{$\surd$} & \multicolumn{1}{c}{$\surd$} & \multicolumn{1}{c|}{$\surd$} & \multicolumn{1}{l|}{1e-4(fix)} & \multicolumn{1}{c|}{41.17} & \multicolumn{1}{c}{0.9834} & \multicolumn{1}{|c}{0.82 M}
\\
\multicolumn{1}{c|}{(f)} & \multicolumn{1}{c}{$\surd$} & \multicolumn{1}{c}{$\surd$} & \multicolumn{1}{c|}{$\surd$} & \multicolumn{1}{l|}{2e-4(fix)} & \multicolumn{1}{c|}{41.27} & \multicolumn{1}{c}{0.9836} & \multicolumn{1}{|c}{0.82 M}
\\
\multicolumn{1}{c|}{(g)} & \multicolumn{1}{c}{$\surd$} & \multicolumn{1}{c}{$\surd$} & \multicolumn{1}{c|}{$\surd$} & \multicolumn{1}{l|}{5e-4(fix)} & \multicolumn{1}{c|}{41.30} & \multicolumn{1}{c}{0.9837} & \multicolumn{1}{|c}{0.82 M}
\\ 
\multicolumn{1}{c|}{OCTUF} & \multicolumn{1}{c}{$\surd$} & \multicolumn{1}{c}{$\surd$} & \multicolumn{1}{c|}{$\surd$} & \multicolumn{1}{c|}{5e-4(warmup)} & \multicolumn{1}{c|}{\textbf{41.34}} & \multicolumn{1}{c}{\textbf{0.9838}} & \multicolumn{1}{|c}{0.82 M}
\\ 
\toprule[1pt] % \hline
\end{tabular}
\vspace{-6pt}
\end{table*}

\subsection{Qualitative Evaluation}

We compare our proposed methods with recent representative CS reconstruction methods. The average PSNR/SSIM reconstruction performances on Set11 dataset \cite{Kulkarni2016ReconNetNR} with respect to five CS ratios are summarized in \cref{tab:set11}. For our OCTUF, we set the iteration number as 10 and set the initial learning rate as $5$$\times 10^{-4}$. To further improve the model performance, we also present a plus version, namely OCTUF$^{+}$, whose iteration number is 16 and initial rate is $2$$\times 10^{-4}$. From \cref{tab:set11},  one can observe that our OCTUF and OCTUF$^+$ outperform all the other competing methods in PSNR and SSIM across all the cases. On average, OCTUF$^+$ outperforms ISTA-Net$^+$ \cite{zhang2018ista}, DPA-Net \cite{sun2020dual}, AMP-Net \cite{zhang2020amp}, MAC-Net \cite{josselyn2020memory}, COAST \cite{you2021coast}, MADUN \cite{song2021memory}, CASNet \cite{chen2022content}, TransCS \cite{shen2022transcs} and FSOINet \cite{chen2022fsoinet} by 3.60 dB, 3.91 dB, 1.25 dB, 3.36 dB, 2.21 dB, 0.51 dB, 0.41 dB, 1.16 dB and 0.29 dB in terms of PSNR on Set11 dataset, respectively. In addition, the average SSIM of OCTUF$^+$ can be improved 0.0389, 0.0280, 0.0113, 0.0381, 0.0202, 0.0013, 0.0015, 0.0073 and 0.0010, respectively. \cref{fig:set11} further show the visual comparisons on challenging images when CS ratio is $25\%$, which can be seen that our OCTUFs can recover much clear edge information than other methods.
% the edge information is much clearer with OCTUFs.

Furthermore, in \cref{tab:urban100}, we compare OCTUFs with other methods on Urban100 dataset \cite{dong2018denoising} that contains more high-resolution images and incorporates more abundant image distributions. It shows that both OCTUF and OCTUF$^+$ achieve a better reconstruction quality at all sampling ratios. \cref{fig:urban100} presents the visual comparisons on challenging images. Our OCTUF and OCTUF$^+$ generate images that are visually pleasant and faithful to the groundtruth. It should be noted that the images on Urban100 dataset do not satisfy a special constraint of MR-CCSNet \cite{fan2022global} that all the image sizes must be divisible by 4, so the performance of MR-CCSNet is only presented on Set11 dataset.

%%%%%%%%%%%%%%%%%%%%%%%%%%%%%%%%%%%%%%%%

\subsection{Ablation Study}

In this part, we conduct ablation studies on Set11 dataset for our OCTUF whose iteration number is 10.

% \subsubsection{Break-down Ablation}
\vspace{5pt}
\noindent \textbf{Break-down Ablation}. We first conduct a break-down ablation experiment in the case of CS ratio $=50\%$ to investigate the effect of each component towards higher performance. The results are listed in \cref{tab:ablation study}. Case (a) is our baseline which contains ResBlock \cite{he2016deep} with a similar number of parameters as OCT module. When we successively apply our FFN and Dual-CA sub-modules respectively, namely Cases (b) and (c), the model achieves 0.71 dB and 2.91 dB improvements. And the model can greatly enhance 3.09 dB gains with little storage place when both sub-modules are used together. We also discuss the effect of the LayerNorm (LN) function in Case (d), which addresses that our OCTUF achieves better performance with the LayerNorm function. Note that without ``LayerNorm'' represents removing all LN from our OCTUF. What is more, we train our models with different learning rates as seen from Cases (e), (f), and (g). ``fix'' denotes that the learning rate is not changed during training, and ``warmup'' denotes that the training strategy is the same with our work as shown in \cref{ssec:exper_detail}.  Our proposed OCTUF has a stable training process with large learning rates and meanwhile, we use the ``warmup'' strategy to improve its anytime performance when training.
\vspace{5pt}

%%%%%%%%%%%88%%%%%%%%%%%%%%%%%%
\begin{table}[t]
\centering
% \small
% \normalsize
%\footnotesize 
\setlength{\abovecaptionskip}{0pt}
\setlength{\belowcaptionskip}{0pt}
\caption{Ablation of Dual-CA sub-module on Set11 dataset \cite{Kulkarni2016ReconNetNR} when CS ratio is $30\%$. ``IF'' denotes the inertial force achieved by the easy way and ``FD'' denotes the enhanced iterative process in the feature domain. The best PSNR(dB) is labeled in bold.} 
% \vspace{-6pt}
\label{tab:attention}
\renewcommand\tabcolsep{4pt}
\begin{tabular}{c|c|c|c|c|c|c|c}
% \hline
\toprule[1pt]
\rowcolor{color3} \multicolumn{1}{c|}{\multirow{1}{*}{Cases}} & \multicolumn{1}{c}{\multirow{1}{*}{FFN}} & \multicolumn{1}{c}{\multirow{1}{*}{GDB}} & \multicolumn{1}{c}{\multirow{1}{*}{IF}} & \multicolumn{1}{c}{\multirow{1}{*}{FD}} & \multicolumn{1}{c}{\multirow{1}{*}{PGCA}} & \multicolumn{1}{c|}{\multirow{1}{*}{ISCA}} & \multicolumn{1}{c}{\multirow{1}{*}{PSNR}}
\\ \hline  
\multicolumn{1}{c|}{(a)} & \multicolumn{1}{c}{$\surd$} & \multicolumn{1}{c}{-} & \multicolumn{1}{c}{-} & \multicolumn{1}{c}{-} & \multicolumn{1}{c}{-} & \multicolumn{1}{c|}{-} & \multicolumn{1}{c}{34.59} 
\\  
\multicolumn{1}{c|}{(b)} & \multicolumn{1}{c}{$\surd$} & \multicolumn{1}{c}{$\surd$} & \multicolumn{1}{c}{-} & \multicolumn{1}{c}{-} & \multicolumn{1}{c}{-} & \multicolumn{1}{c|}{-} & \multicolumn{1}{c}{35.93} 
\\
% \multicolumn{1}{c|}{(c)} & \multicolumn{1}{c}{$\surd$} & \multicolumn{1}{c}{$\surd$} & \multicolumn{1}{c}{$\surd$} & \multicolumn{1}{c}{-} & \multicolumn{1}{c}{-} & \multicolumn{1}{c|}{-} & \multicolumn{1}{c|}{} & \multicolumn{1}{c}{} & \multicolumn{1}{|c}{}
% \\ 
\multicolumn{1}{c|}{(c)} & \multicolumn{1}{c}{$\surd$} & \multicolumn{1}{c}{$\surd$} & \multicolumn{1}{c}{-} & \multicolumn{1}{c}{$\surd$} & \multicolumn{1}{c}{-} & \multicolumn{1}{c|}{-} & \multicolumn{1}{c}{36.82} 
\\ 
\multicolumn{1}{c|}{(d)} & \multicolumn{1}{c}{$\surd$} & \multicolumn{1}{c}{$\surd$} & \multicolumn{1}{c}{$\surd$} & \multicolumn{1}{c}{$\surd$} & \multicolumn{1}{c}{-} & \multicolumn{1}{c|}{-} & \multicolumn{1}{c}{36.83} 
\\ 
\multicolumn{1}{c|}{(e)} & \multicolumn{1}{c}{$\surd$} & \multicolumn{1}{c}{$\surd$} & \multicolumn{1}{c}{-} & \multicolumn{1}{c}{$\surd$} & \multicolumn{1}{c}{-} & \multicolumn{1}{c|}{$\surd$} & \multicolumn{1}{c}{37.13} 
\\
\multicolumn{1}{c|}{(f)} & \multicolumn{1}{c}{$\surd$} & \multicolumn{1}{c}{-} & \multicolumn{1}{c}{-} & \multicolumn{1}{c}{$\surd$} & \multicolumn{1}{c}{$\surd$} & \multicolumn{1}{c|}{-} & \multicolumn{1}{c}{37.08}
\\ 
\multicolumn{1}{c|}{OCTUF} & \multicolumn{1}{c}{$\surd$} & \multicolumn{1}{c}{-} & \multicolumn{1}{c}{-} & \multicolumn{1}{c}{$\surd$} & \multicolumn{1}{c}{$\surd$} & \multicolumn{1}{c|}{$\surd$} & \multicolumn{1}{c}{\textbf{37.21}} 
\\ 
\toprule[1pt] % \hline
\end{tabular}
\vspace{-16pt}
\end{table}
%%%%%%%%%%%%%%%%%%%%%%%%%%%%%%%%%%%%%%%%

\begin{figure}[t]
\centering
\begin{minipage}[t]{0.24\linewidth}
\centering
% \small
\includegraphics[width=1\columnwidth]{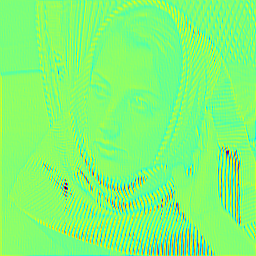}\\ w/o Dual-CA \\
\end{minipage}
\begin{minipage}[t]{0.24\linewidth}
\centering
\includegraphics[width=1\columnwidth]{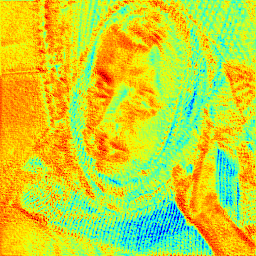}\\ w/o PGCA \\
\end{minipage}
\begin{minipage}[t]{0.24\linewidth}
\centering
\includegraphics[width=1\columnwidth]{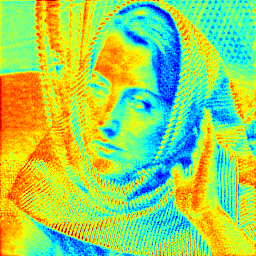}\\ w/o ISCA \\
\end{minipage}
\begin{minipage}[t]{0.24\linewidth}
\centering
\includegraphics[width=1\columnwidth]{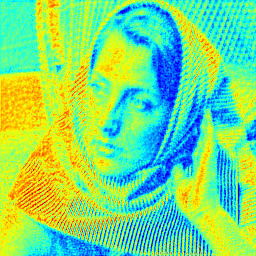}\\ OCTUF \\
\end{minipage}
\centering
\vspace{-3pt}
\caption{Visual analysis of the feature map in the fifth iteration of our proposed OCTUF.
% including the cases: (1) OCTUF w/o Dual-CA; (2) OCTUF w/o PGCA; (3) OCTUF w/o ISCA; (4) OCTUF. 
It shows that both ISCA and PGCA blocks pay more high-fidelity attention to details.
}
\vspace{-6pt}
\label{heatmap} 
\end{figure}

% \subsubsection{Analysis of Dual Cross-channel Attention}
% \vspace{5pt}
\noindent \textbf{Dual Cross Attention}. We also do elaborate ablation experiments on the components of Dual-CA sub-module in the case of CS ratio $=30\%$ in \cref{tab:attention}, where ``IF'' denotes the inertial force achieved by a simple way similar to \cref{eq:r} and ``FD'' represents that the overall iteration process is achieved in feature domain. Case (b) achieves 1.34 dB improvement compared with Case (a), which proves the superiority of DUN compared with the structure that only contains a neural network. Then, the performance can continuously improve by 0.89 dB after using ``FD'' shown in Case (c). We also conduct fine contrast experiments for the inertial force in Cases (c), (d), and (e), demonstrating that our ISCA block can more fully play the role of inertia force. And as shown in Cases (e)(f), applying PGCA and ISCA blocks can get better performance. Our proposed Dual-CA sub-module takes into account the gradient descent algorithm and the inertial force, allocates the different channels reasonably, and gives full play to the structural characteristics.
% without increasing the number of parameters as much as possible. 
\cref{tab:attention} should be also noted that ``GDB'' is included in ``PGCA'' so it is not selected when ``PGCA'' is selected. Moreover, to intuitively show the advantages of Dual-CA sub-module, we visualize the feature map in the fifth iteration for four cases. The result in \cref{heatmap} presents that both our ISCA and PGCA blocks pay more high-fidelity attention to the detailed contents and structural textures.

% \subsubsection{Analysis of Proximal Mapping Feed-Forward Network}
\vspace{5pt}
\noindent \textbf{Feed-Forward Network}. As shown in \cref{fig:each_stage} (e), our proposed Feed-Forward Network (FFN) sub-module consists of two groups of LayerNorm and Feed-Forward Block (FFB) with a global skip connection.
% , which is different from the traditional Transformers that contain only one group \cite{cai2022degradation,cai2022mask}.
We do ablations to investigate the effects of the group number and LayerNorm (LN) number in \cref{tab:pmffn}. ``Baseline'' is the same setting with Case (c) of \cref{tab:ablation study}, ``LN+FFB'' denotes one group, and ``LN+$2 \times$FFB'' denotes that Norm is only added to the first group. Therefore, as can be seen from the table, our proposed method achieves the best performance.

%%%%%%%%%%%%%%%%%%%%%%%%%%%%%%%%%%%%%%%%
%%%%%%%%%%%%%%%%%%%%%%%%%%%%%%%%%%%%%%%%
% %%%%%%%%%%%%%%%%%%%%%%%%%%%%%%%%%%%%%
\begin{table}[!t]
% \vspace{-12pt}
\centering
\small
% \normalsize
% \footnotesize 
\setlength{\abovecaptionskip}{0pt}
\setlength{\belowcaptionskip}{0pt}
\caption{Ablation of Feed-Forward Network on Set11 dataset \cite{Kulkarni2016ReconNetNR} when CS ratio $=30\%$. The best performance is labeled in bold.} 
% \vspace{-3pt}
\label{tab:pmffn}
\renewcommand\tabcolsep{6pt}
\begin{tabular}{ccccc}
\toprule
% \hline
\rowcolor{color3} \multicolumn{1}{c|}{Method} & \multicolumn{1}{c}{Baseline} & \multicolumn{1}{c}{LN+FFB} & \multicolumn{1}{c}{LN+$2 \times$FFB} & \multicolumn{1}{c}{Ours}
\\ \hline  
\multicolumn{1}{c|}{PSNR(dB)}  & \multicolumn{1}{c}{37.05} & \multicolumn{1}{c}{37.04} & \multicolumn{1}{c}{37.12} & \multicolumn{1}{c}{\textbf{37.21}}
\\
\multicolumn{1}{c|}{SSIM}  & \multicolumn{1}{c}{0.9663} & \multicolumn{1}{c}{0.9664} & \multicolumn{1}{c}{0.9672} & \multicolumn{1}{c}{\textbf{0.9673}}
\\
\multicolumn{1}{c|}{Parameters}  & \multicolumn{1}{c}{0.61 M} & \multicolumn{1}{c}{0.52 M} & \multicolumn{1}{c}{0.61 M} & \multicolumn{1}{c}{0.61 M}
\\
\toprule
% \hline
\end{tabular}
\vspace{-16pt}
\end{table}
%%%%%%%%%%%%%%%%%%%%%%%%%%%%%%%%%%%%%%%%

\subsection{Complexity Analysis}

The computation cost and model size are important in many practical applications. \cref{tab:complexity} provides the comparisons of the parameters, the model size, and FLOPs for reconstructing a $256$$ \times 256$ image when CS ratio is $10\%$. CASNet \cite{chen2022content} designs a complex sampling network and reconstruction network, which has a large number of parameters and computational overhead. Our OCTUFs have the same parameters/cost for the sampling process with MADUN and FSOINet, but use fewer parameters and less computation burden to produce much sharper recovered images.

% %%%%%%%%%%%%%%%%%%%%%%%%%%%%%%%%%%%%%
\begin{table}[!t]
% \vspace{-12pt}
\centering
\small
% \normalsize
% \footnotesize 
\setlength{\abovecaptionskip}{0pt}
\setlength{\belowcaptionskip}{0pt}
\caption{Comparison of the parameters, the model size and FLOPs for reconstructing a $256$$ \times 256$ image in the case of CS ratio $=10\%$. The best performance is labeled in bold.} 
% \vspace{-3pt}
\label{tab:complexity}
\renewcommand\tabcolsep{3pt}
\begin{tabular}{cccccc}
\toprule
% \hline
\rowcolor{color3} \multicolumn{1}{c|}{Method} & \multicolumn{1}{c}{MADUN} & \multicolumn{1}{c}{CASNet} & \multicolumn{1}{c}{FSOINet} & \multicolumn{1}{c}{OCTUF} & \multicolumn{1}{c}{OCTUF$^{+}$}
\\ \hline  
\multicolumn{1}{c|}{Params.(M)}  & \multicolumn{1}{c}{3.14} & \multicolumn{1}{c}{16.90} & \multicolumn{1}{c}{0.64} & \multicolumn{1}{c}{\textbf{0.40}} & \multicolumn{1}{c}{0.58}
\\
\multicolumn{1}{c|}{Size(MB)}  & \multicolumn{1}{c}{11.9} & \multicolumn{1}{c}{66.3} & \multicolumn{1}{c}{7.8} & \multicolumn{1}{c}{\textbf{5.2}} & \multicolumn{1}{c}{7.5}
\\
\multicolumn{1}{c|}{FLOPs(G)}  & \multicolumn{1}{c}{419.2} & \multicolumn{1}{c}{13391.5} & \multicolumn{1}{c}{266.6} & \multicolumn{1}{c}{\textbf{189.3}} & \multicolumn{1}{c}{294.6}
\\
\toprule
% \hline
\end{tabular}
\vspace{-6pt}
\end{table}
% %%%%%%%%%%%%%%%%%%%%%%%%%%%%%%%%%%%%%
%%%%%%%%%%%%%%%%%%%%%%%%%%%%%%%%%%%%%%%%
\begin{figure}[t]
\centering
\begin{minipage}[t]{0.48\linewidth}
\centering
% \small
\includegraphics[width=1\columnwidth]{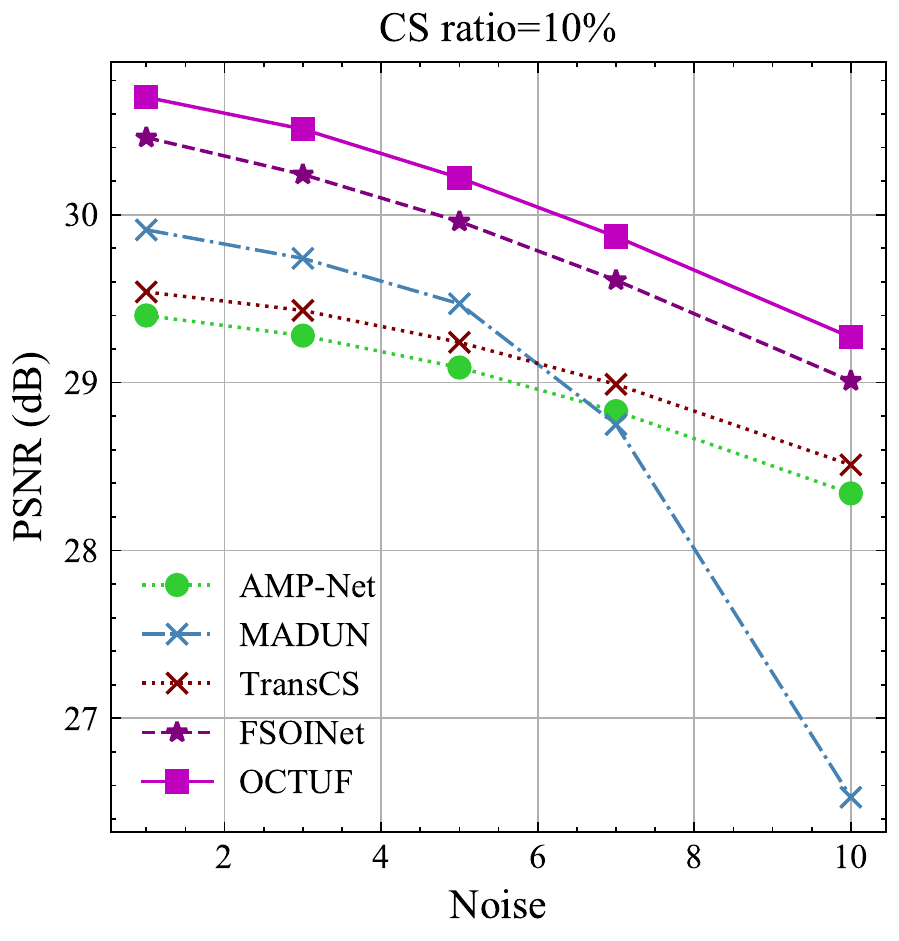} \\
\end{minipage}
\begin{minipage}[t]{0.48\linewidth}
\centering
\includegraphics[width=1\columnwidth]{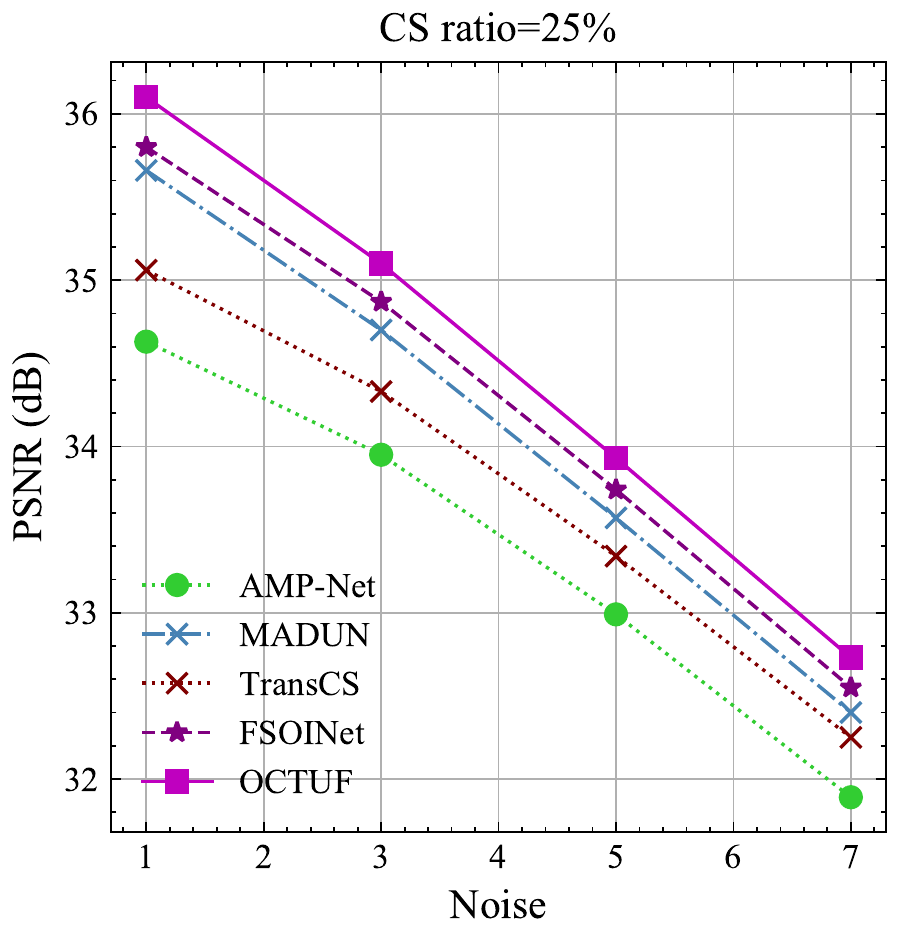} \\
\end{minipage}
\vspace{-3pt}
\caption{Comparison of robustness to Gaussian noise. 
% between various methods.
}
\vspace{-15pt}
\label{fig:noise} 
\end{figure}
%%%%%%%%%%%%%%%%%%%%%%%%%%%%%%%%%%%%%%%%

\subsection{Sensitivity to Noise}

In the real application, the imaging model may be affected by noise and currently, no real open dataset is suitable for such CS reconstruction methods. So to test the robustness of our designed OCTUF, we first add the Gaussian noise with different noise levels to the original images for Set11 dataset. Then, OCTUF and the other methods take the noisy images as input, and sample and recover the reconstruction images when CS ratios are $10\%$ and $25\%$. \cref{fig:noise} shows the plots of PSNR values of all methods versus various standard variances noise. It can be seen that our OCTUF possesses strong robustness to noise corruption.

\section{Conclusion}

In this paper, we propose a novel optimization-inspired cross-attention Transformer (OCT) module as an iteration, leading to a lightweight OCT-based unfolding framework (OCTUF) for CS. Specifically, we present a Dual Cross Attention (Dual-CA) sub-module, which contains an inertia-supplied cross attention (ISCA) block and a projection-guided cross attention (PGCA) block as the projection step in iterative optimization. ISCA block precisely helps to achieve a good convergence by introducing a feature-level inertial term. PGCA block utilizes a cross attention mechanism to fuse the gradient descent term and the inertial term while ensuring the maximum information flow. Extensive experiments show that our OCTUF achieves superior performance compared to state-of-the-art methods with lower complexity. In the future, we will extend our OCTUF to other image inverse problems and video applications.

%%%%%%%%% REFERENCES
{\small
\bibliographystyle{ieee_fullname}
\bibliography{egbib}
}

\end{document}